\documentclass{article}
\usepackage{spconf,amsmath,graphicx}

\usepackage[numbers,sort&compress]{natbib}
\usepackage{amssymb}
\usepackage{booktabs}
\usepackage{enumitem}
\usepackage{stfloats} 
\usepackage{graphicx}
\usepackage{color}

\newcommand{\yuzhao}[1]{}
\title{learnable hypergraph Laplacian for hypergraph learning}
\name{Jiying Zhang$^{1, \star}$ \quad Yuzhao Chen$^{1, \star}$ \quad Xi Xiao$^{1, 2, \dagger}$
\quad Runiu Lu$^{1}$ \quad Shu-Tao Xia$^{1, 2}$ 
\thanks{$^{\star}$The first two authors contribute equally to this work.} \thanks{$^{\dagger}$Corresponding author: Xi Xiao}}

\address{$^{1}$Tsinghua Shenzhen International Graduate School, Tsinghua University, Shenzhen, China \\
$^{2}$ Peng Cheng Laboratory, Shenzhen, China\\
\tt\small \{zhangjiy20, chen-yz19\}@mails.tsinghua.edu.cn, \tt\small hugw@sziit.edu.cn \\  
\tt\small \{xiaox,lurn,xiast\}@sz.tsinghua.edu.cn
}
\begin{document}
%

\maketitle

\begin{abstract}
HyperGraph Convolutional Neural Networks (HGCNNs) have demonstrated their potential in modeling high-order relations preserved in graph structured data. 
However, most existing
convolution filters
are localized and determined by
the pre-defined initial hypergraph topology,
neglecting to explore implicit and long-range relations in real-world data.
In this paper, we propose the \emph{first} learning-based method tailored for constructing adaptive hypergraph structure,
termed HypERgrAph Laplacian aDaptor (HERALD), 
which serves as a generic plug-in-play module for improving the representational power of HGCNNs.
Specifically, HERALD adaptively optimizes the adjacency relationship between hypernodes and hyperedges in an end-to-end manner and thus the task-aware hypergraph is learned.
Furthermore, HERALD employs the self-attention mechanism to capture the non-local paired-nodes relation.
Extensive experiments on various popular hypergraph datasets for node classification and graph classification tasks demonstrate that our approach obtains consistent and considerable performance enhancement, proving its effectiveness and generalization ability.
\end{abstract}


\section{Introduction}
\label{sec:intro}
In recent years, Graph Convolutional Neural Networks (GCNNs) develop rapidly~\cite{kipf2016semi,defferrard2016convolutional},
due to their ability of processing 
graph structured data, such as social networks~\cite{yanardag2015deep}, citation networks~\cite{sen2008collective} and biomedical networks~\cite{yue2020graph,torng2019graph}. 
GCNNs have shown superiority on graph representation learning compared with traditional neural networks which are used to process regular data.
\par Many graph neural networks have been developed to model ordinary graph whose edge connects exactly two vertices. 
At the meantime, more and more researchers noted that the data structure in real practice could be beyond pair connections, and intuitive pairwise connections among nodes are usually insufficient for capturing  higher-order relationships. 

Consequently, a new research area aiming at establishing convolutional networks on hypergraphs attracts a surge of attention recently. 
Hypergraph is a generalization of the regular graph, whose edge could join any number of vertices, and thus possess a more powerful capability of 
modeling complex relationship preserved in the real-world data \cite{konstantinova2001application,maruyama2001learning,tu2018structural}. 
For example, in a co-citation relationship~\cite{hypergcn}, papers act as hypernodes, and citation relationships become hyperedges.  
Existing representative works in literature include HGNN~\cite{HGNN}, HyperGCN~\cite{hypergcn}, HNHN~\cite{HNHN}, 
which have  designed basic network architecture for hypergraph learning.
However, a major concern is that these methods are built upon an intrinsic hypergraph and 
not capable of dynamically optimizing the hypergraph topology.
The topology plays an essential role in the message passing across nodes, and its quality could make a significant impact on the performance of the trained model~\cite{agcn}.  
Another barrier that limits the representational power of HGCNNs is that they
only aggregate message of hypernodes in a localized range, while neglecting to explore information about
long-range relations.

Several works have made their attempts to remedy these issues.
DHSL \cite{DHSL} 
proposes to use the initial raw graph 
to update the hypergraph structure, but it fails to capture high-order relations among features.  Also, the  optimization algorithm in DHSL could be expensive cost and unable to incorporate with convolutional networks as well.
DHGNN \cite{DHGNN} manages to capture local and global feature, but 
the adopted K-NN method~\cite{altman1992introduction} leads the 
graph structure to a k-uniform hypergraph and lost its flexibility.
AGCN \cite{agcn} proposes a spectral graph convolution network  that can optimize graph adjacent matrix during training. 
Unfortunately, 
it can not be naturally extended to hypergraph spectral learning for which is based on the incidence matrix. 

In a hypergraph, incidence matrix represents its topology by recording the connectioins and connection strength between hypernodes and hyperedges. 
As an intuition, one could parameterize such a matrix and involve it into the end-to-end training process of the network.
To this end, 
we propose a noval hypergraph laplacian adaptor (HERALD), 
the \emph{first}  fully  learnable module designed
for adaptively optimizing the hypergraph structure. 
Specifically, 
HERALD takes the node features and the pre-defined hypergraph laplacian as input, 
and then
constructs the parameterized distance matrix between nodes and hyperedges, which empowers the automated updating of the hypergraph laplacian and thus the topology is adapted for the down-stream task.
Notably, HERALD employs the self-attention mechanism to model the non-local paired-nodes relation for embedding  global property of the hypergraph into  the learned topology.
In the experiments, to evaluate the performance of the proposed module, 
we have conducted experiments for 
hypergraph classification tasks on  node-level and graph-level.
Our main contributions are summarized as below:
\begin{enumerate}[leftmargin=*]
    \item We propose a generic and plug-in-play module, termed HypERgrAph Laplacian aDaptor (HERALD),
    for automated adapting the hypergraph topology to the specific down-stream task. It is the first learnable module that can update hypergraph structure dynamically.
    \item HERALD adopts the self-attention mechanism to capture global information on the hypergraph and the parameterized distance matrix is built, which empowers the learning of the topology in an end-to-end manner.
    \item We have conducted extensive experiments on node classification and graph classification tasks, 
    and the results show that consistent and considerable performance improvement is obtained, which verifies the effectiveness of the proposed approach.
    
\end{enumerate}
\par The rest of this paper is organized as follows.
Section 2 introduces the basic formulation of the spectral convolution of hypergraph and some common notations.
Section 3 gives details about the proposed module HERALD. 
Section 4 shows 
the experiment results
and Section 5 concludes this paper.


\section{learning of hypergraph Laplacian}
\label{sec:format}
\textbf{Notation.} Let $\mathcal{G}=(\mathcal{V,E})$  represents the input hypergraph  with hypernode set of $\mathcal V$ and hyperedge set of $\mathcal E$. 
Hyperedge weights are assigned by a diagonal matrix $\mathbf W\in \mathbb{R}^{|\mathcal{E}|\times |\mathcal{E}|}$.
The structure of hypergraph $\mathcal G$ can be denoted by a incidence matrix $\mathbf{H}\in \mathbb{R}^{|\mathcal{V}|\times |\mathcal{E}|}$ with each entry of $h(v,e)$, which equals 1 when $e$ is incident with $v$ and 0 otherwise. 
The degree of hypernode and hyperedge are defined as $d(v)=\sum_{e\in \mathcal{E}}w(e)h(v,e)$ and $\delta
(e)=\sum_{v\in\mathcal V}h(v,e)$ which can be denoted by diagonal matrixes $\mathbf{D}_v\in \mathbb{R}^{|\mathcal{V}|\times |\mathcal{V}|} $ and $\mathbf{D}_e\in \mathbb{R}^{|\mathcal{E}|\times |\mathcal{E}|}$.


\textbf{Hypergraph Spectral Convolution}. 
The normalized hypergraph laplacian matrix $\mathbf{L}$ is given by:
\begin{equation}
\label{laplacian}
\mathbf L=\mathbf{I}-\mathbf{D}_v^{-1/2}\mathbf{HW}\mathbf{D}_e^{-1}\mathbf{H}^\top\mathbf{D}_v^{-1/2},
\end{equation}
which represents the connection strength of every pair of nodes.
Since $\mathbf L$ is a symmetric positive semi-definite matrix, 
then its spectral decomposition is
$\mathbf L=\mathbf U\Lambda\mathbf U^\top$, where $\mathbf U=\{u_1,u_2,...,u_{n}\}$ is the set of eigenvectors  and $\Lambda=diag(\lambda_1,\lambda_2,...,\lambda_{n})$ contains the corresponding eigenvalues. The Fourier transform is then defined as $\hat x=\mathbf Ux$. 
The spectral convolution of signal $\mathbf{x}$ and filter $g$ can be denoted as:
\begin{equation}
    \mathbf{g}\star \mathbf{x}=\mathbf Ug(\Lambda)\mathbf U^\top \mathbf{x} \label{conv0},
\end{equation}
here, $g(\Lambda)=diag(g(\lambda_1),...,g(\lambda_{n}))$ is a function of the Fourier coefficients.
Following  spectral graph convolution\cite{defferrard2016convolutional}, 
we can 
get the polynomial parametrization of filter $g$:
\begin{equation}
    g_\theta(\Lambda)=\sum\nolimits_{k=0}^{K-1}\theta_k\Lambda^k \label{kernel},
\end{equation}
Thus, the convolution 
calculation
for hypergraph results in:
\begin{equation}
    \mathbf{g}\star \mathbf x=\mathbf U\sum\nolimits_{k=0}^{K-1}\theta_k\Lambda^k\mathbf U^\top \mathbf{x}=\sum\nolimits_{k=0}^{K-1}\theta_k \mathbf L^k \mathbf x \label{conv1}.
\end{equation}


\section{Methodology}
\label{sec:pagestyle}
Although we can get the spectral convolution in Eq.\eqref{conv1}, it is noticed that the convolution kernel  Eq.\eqref{kernel} is only a $\textit{K}$-localized kernel, 
which aggregates 
$\textit{K}$-hop nodes 
to the farthest per iteration
and thus 
restricts the flexibility of kernel.
Furthermore, the initial pre-defined 
hypergraph structure is not always the optimal one for 
the specific down-stream learning task.
Notably, it's somehow proved that GCNs which are  $\textit{K}$-localized and topology-fixed actually simulate a  polynomial filter with fixed coefficients~\cite{li2018deeper,chen2020simple}.
As a result, existing techniques~\cite{HGNN,hypergcn,HNHN}
might neglect the modeling of  non-local information and fail in obtaining high-quality hypergraph embeddings  as well.

In order to improve the representaional power of HyperGraph convolutional Neural Networks (HGCNNs), we 
propose the HypERgrAph Laplacian aDaptor (HERALD) to dynamically optimize the hypergraph structure (a.k.a. adapt the hypergraph topology to the specific down-stream task).
The HERALD focuses the directly learning of the filter $g$ to make a more flexible spectral convolution kernel and captures the global information that is beneficial to the adapting of the hypergraph topology.
Based on Eq.\eqref{kernel}, the new filter is defined as:
\begin{equation}
    g_\theta(\Lambda)=\sum\nolimits_{k=0}^{K-1}(\mathcal{T}(\mathbf{L,X,\Theta}))^k,
\end{equation}
where $\mathbf{L}$ and $\mathbf{X}$ are the original Laplacian matrix and node features respectively. 
$\mathbf\Theta$ is 
the parameter and  $\mathcal T$ is the 
learnable function that outputs the spectrum of updated Laplacian $\mathbf {\Tilde{L}}$.
Then, the spectral convolution will become:
\begin{equation}
    \mathbf Y=  \mathbf{g}\star  \mathbf x=\mathbf U\sum\nolimits_{k=0}^{K-1}(\mathcal{T}(\mathbf{L,X,\Theta}))^k\mathbf U^\top \mathbf{x} \label{conv_new}.
\end{equation}

\textbf{Learnable Incidence Matrix}. In order to learn a suitable hypergraph structure, 
HERALD takes the node features and the pre-defined togology to construct a parametrized incidence matrix. To be specific, 
given original incidence matrix $\mathbf H\in\mathbb{R}^{|\mathcal{V}|\times |\mathcal{E}|}$ and node features $\mathbf X=\{x_1;...;x_{|\mathcal V|}\}\in \mathbb{R}^{|\mathcal{V}|\times d}$ , we first get the hyperedge features by averaging the features of incident hypernodes.
\begin{equation}
    x_{e_i}=\frac{1}{|e_i|}\sum_{v\in e_i}x_v,
\end{equation}
where
$|e_i|$ denotes the number of nodes in hyperedge $e_i$. Then we use a linear transformation to obtain the transformed hyperedge feature:
\begin{equation}
    x_{e_i}=\mathbf{W}_e^Tx_{e_i},
\end{equation}
where $\mathbf{W}_e\in \mathbb{R}^{d\times h}$ is the learnable parameter.
Next, in order to enhance the representational power of the convolution kernel, 
HERALD adopts the self-attention mechanism~\cite{vaswani2017attention}
to encode the non-local relations between paired nodes into the updated node features $\mathbf{X}$.
That is to say, the enhanced node features are formulated as:
\begin{equation}
    x_i=\sum_{1\leq j \leq k}\alpha_{ij}\mathbf{W}_v^Tx_j,
\end{equation}
where $\mathbf{W}_v$ is set to have the same dimension as $\mathbf{W}_e$, 
and the attention weights are calculated by:
\begin{equation}
    \gamma_{ij}=(\mathbf{W}_v^Tx_i)^\top(\mathbf{W}_v^Tx_j), \; 1\leq i,j\leq|\mathcal{V}|,
\end{equation}
\begin{equation}
    \alpha_{i,j}=\frac{exp(\gamma_{ij})}{\sum_{1\leq k\leq |\mathcal{V}|}exp(\gamma_{ik})}
\end{equation} 
With the generated node features $\{x_1,...,x_{|\mathcal{V}|}\}$ and hyperedge features $\{x_{e_1},...,x_{e_{|\mathcal{E}|}}\}$, 
we calculate the Hardamard power (element-wise power)~\cite{sagnn} of each pair of 
hypernode and hyperedge. 
And then we obtain the pseudo-euclidean distance matrix of hyperedge and hypernode after a linear transformation: 
\begin{align}
    d_{ij}=\mathbf{W}_s^T(x_i-x_{e_j})^{\circ 2},  \; 1\leq i\leq |\mathcal{V}|, 1\leq j \leq  |\mathcal{E}|,    
\end{align}
where  $\mathbf{W}_s\in \mathbb{R}^{h\times 1}$ and $(\cdot)^{{\circ 2}}$ denotes the Hardamard power. 
Finally, the soft assignment matrix H is constructed by further parameterizing the generated distance matrix with a  Gaussian kernel, in which each element represents the probability that the paired node-hyperedge is connected:
\begin{equation}
\Tilde{\mathbf H}_{i,j}=exp(-d_{ij}/2\sigma^2) \label{G},
\end{equation}
where the hyper-parameter of $\sigma$ is fixed to 20 in our experiments.
Based on the parametrized $\mathbf{\Tilde{H}}$ and Eq.\eqref{laplacian}, HERALD outputs the the hypergraph Laplacian $\mathbf {\Tilde{L}}$. 

\textbf{Residual connect}. 
In fact, if we learn the hypergraph topology from scratch,
it may 
spend expensive cost for optimization
converge 
due to the lacking prior knowledge 
about a proper initialization on the parameters.
To this end, we
reuse the intrinsic graph structure to accelerate the training and increase the 
training
stability. 
To simplify the notation, we denote  $\mathbf{D}_v^{-1/2}\mathbf{HW}\mathbf{D}_e^{-1}\mathbf{H}^\top\mathbf{D}_v^{-1/2}$ as $\mathbf{N}$ and thus the normalized hypergraph laplacian matrix $\mathbf{L}=\mathbf{I}-\mathbf{N}$.
Formally, 
we assume
that 
the optimal $\hat{\mathbf N}$ is 
slightly shifted away
from the original $ \mathbf{N}$:
\begin{equation}
    \hat{\mathbf{N}}=(1-a) \mathbf{N}+a \mathbf{N}_{res} \label{residue},
\end{equation}
where $a$ is the hyper-paremeter that controls the updating strength of the topology.  
From this respect,
the HERALD module actually learns the residual $\mathbf N_{\text{res}}$ rather than $\hat{\mathbf{N}}$.     

To sum up, the learning complexity of HERALD module is $\mathcal O(d_{i}h)$ ($d_i$ is the dimension of the input features at  layer $i$) with introduced parameters $\{\mathbf{W}_v,\mathbf{W}_e,\mathbf{W}_s\}$,  independent of the input hypergraph size and node degrees.
Finally, the  algorithm of HERALD is summarized as follow:
\\
\\
\scalebox{0.95}{
\begin{tabular}{l}
  
    \toprule
    \textbf{Algorithm 1} HERALD module \\
    \midrule
       Input: \textbf{Hypergraph Features} $\mathbf{X}=\{\mathbf{x_i}\},\mathbf{N}=\mathbf{I-L}$;\\
       $\mathbf{Parameter}~ a,\mathbf{W_v,W_e,W_s}$\\
     1: $\Tilde{\mathbf H}\gets$ Eq.(7-13) \\
     2: $\mathbf{N}_{res}=\Tilde{\mathbf{D}}_v^{-1/2}\Tilde{\mathbf{H}}\Tilde{\mathbf{W}}\Tilde{\mathbf{D}}_e^{-1}\Tilde{\mathbf{H}}^\top\Tilde{\mathbf{D}}_v^{-1/2}$ \\
     3: $\hat{\mathbf N}=(1-a)\mathbf N + a\mathbf{N}_{res}$ \\
     4: $\mathbf{\Tilde{L}=I-\mathbf{\hat{N}}}$ \\
     5: \textbf{Return} $\mathbf{\Tilde{L}}$ \\
    \bottomrule
\end{tabular}
}

\section{experiments}
\label{sec:typestyle}
Node classification and graph classification are two natural settings of hypergraph learning. Most of hypernode classification problems are semi-supervised learning, where the goal is to assign labels to unlabelled nodes in hypergraph and only a little amount labelled data is accessible~\cite{zhu2009introduction,HGNN}.
While many spectral convolution approaches have been proposed to solve node classification problems~\cite{HGNN,hypergcn,HNHN},  however, there exists no previous works conducting on  the hypergraph classification to the best of our knowledge.
In the experiments, we evaluate the proposed HERALD on both these tasks, 
and we select the representative model of HGNN~\cite{HGNN} to work as the evaluation backbone.
We have also conducted many experiments to compare our method with the vanilla GCN~\cite{kipf2016semi}.

\begin{table*}[htbp]
\centering
\caption{The results on the task of hypergraph classification. We report the average test accuracy and its standard deviation under the  10-fold cross-validation.}
\vspace{1mm}
\scalebox{0.9}{
    \begin{tabular}{c|c|c|c|c|c|c}
        \toprule
              Datasets & MUTAG & PTC & IMDB-B & PROTEINS & NCI1 & COLLAB \\
        \midrule
             \# layers  & 4 & 2 & 3 & 3 & 2 & 3 \\ 
             GCN~\cite{kipf2016semi} & 68.60 $\pm$ 5.6 & 65.41 $\pm$ 3.7 & 53.00 $\pm$ 1.9 & 68.29 $\pm$ 3.7 & 58.08 $\pm$ 1.1 & 52.69 $\pm$ 0.5 \\
             HGNN~\cite{HGNN} & 69.12 $\pm$ 6.2 & 66.56 $\pm$ 4.7 & 55.20 $\pm$ 3.7& 68.38 $\pm$ 3.8 & 58.32 $\pm$ 1.3 & 55.61 $\pm$ 2.5 \\
             HGNN + HERALD & \textbf{71.23 $\pm$ 9.0} & \textbf{67.75 $\pm$ 5.9} & \textbf{58.20 $\pm$ 5.5} & \textbf{68.64 $\pm$ 3.4} & \textbf{58.37 $\pm$ 1.4} & \textbf{55.74 $\pm$ 2.3} \\
        \bottomrule
    \end{tabular}
    }
     \vspace{-4mm}
    \label{graph classification}    
\end{table*}

\subsection{Hypernode Classification}
\label{ssec:node_subhead}
\textbf{Datasets.} In this experiment, 
we employ two hypergraph datasets~\cite{hypergcn}, one is the co-citation relationship of Cora~\cite{sen2008collective}, the other is the co-authorship of it.
The co-authorship data consists of a collection of the papers with their authors and the
co-citation data consists of a collection of the papers and their citation relationship. 
The details of the datasets are shown in Table \ref{datasets}. 

\begin{table}[t]
\centering
\caption{The brief introduction about the node classification datasets used in our work. }
\vspace{1mm}
\scalebox{0.9}{
    \begin{tabular}{c|c|c}
        \toprule
            Dataset & Cora & Cora \\
          & (co-citation) & (co-authorship) \\
        \midrule
         \# hypernodes, $|\mathcal V|$ & 2708 & 2708 \\
         \# hyperedges, $|\mathcal E|$ & 1579 & 1072 \\
         \# classes  & 7 & 7 \\
        \bottomrule
    \end{tabular}
    }
    \label{datasets}    
    \vspace{-4mm}
\end{table}

\textbf{Experimental Setup}.
\label{node setup}
We use the backbone of a  $3$-layer HGNN as the evaluation baseline. 
Since different datasets have different hypergraph structure, we perform random search on the hyper-parameters 
and report the case giving the best accuracy on validation set. 
To evaluate the proposed module, we 
add the HERALD module to the latter two layers in HGNN. 
That is to say,  in the $l$-th layer (where $2\leq l \leq 3$),  HERALD generates $\mathbf{\hat{N}}^{(l)}$,
and the feature updating function is given by $\mathbf X^{(l+1)}=\beta(\mathbf{\hat{N}}^{(l)}\mathbf{X}^{(l)}\mathbf{W})$, where $\beta$ is the non-linear activation function. 
For the hyper-parameter of $a$, we set $a=1-0.9*(\text{cos}(\pi (l-1)/10)+1)/2$ for gradually increase the updating strength of the task-specific adapted topology. 
We also add a loss regularizer of $||\mathbf N-\mathbf{N}_{res}||_2$ to make the training more stable, of which the loss weight is fixed to $0.1$. 

\textbf{Fast-HERALD}.  For constructing a more cost-friendly method, we also propose a variant of the usage of HERALD, named Fast-HERALD.
It  constructs $\hat{\mathbf{N}}$ at the beginning HGNN layer and reuses it in the process of feature updating for all the rest layers. 
Since $\hat{\mathbf{N}}$ is shared on each layer, it can reduce the number of parameters and increase the training speed.

\begin{table}[t]
\centering
\caption{The results on the task of node classification. We report the average test accuracy and its standard deviation of test accuracy under 10 runs with different random seeds.}
\vspace{1mm}
\scalebox{0.9}{
    \begin{tabular}{c|c|c}
        \toprule
            Method & Cora & Cora \\
          & (co-citation) & (co-authorship) \\
        \midrule
         HGNN~\cite{HGNN} & 48.23 $\pm$ 0.2 & 69.21 $ \pm$ 0.3 \\
         HGNN + HERALD  & \textbf{57.31  $\pm$ 0.2} &70.05  $\pm$  0.3 \\
        HGNN + FastHERALD & 57.27 $\pm$ 0.3 &  \textbf{70.16 $\pm$ 0.4} \\
        \bottomrule
    \end{tabular}}
    \vspace{-4mm} 
    \label{node results}    
\end{table}

\textbf{Results and Disscussion}. 
The results of experiments for hypernode classification are given in Table \ref{node results}. It is observed that HERALD module consistently improves the testing accuracy for all the cases. 
It gains 0.84$\%$ improvement on Co-authorship Cora dataset while achieving a remarkable 9.08$\%$ increase on Co-citation Cora dataset. We also notice the FastHERALD get the best result on co-authorship Cora. 
The results show that our proposed module can significantly improve the performance of hypergraph convolutional network by adapting the topology to the downstream task.
\vspace{-1mm}

\subsection{Hypergraph Classification}
\label{ssec:graph_subhead}

\textbf{Datasets}. The task in this experiment is graph classification. The datasets we use include six graph classification benchmarks: four bioinformatics datasets (MUTAG, PTC, NCI1, PROTEINS) and two social network datasets (IMDB-BINARY, COLLAB)~\cite{yanardag2015deep}. However, all those datasets are ordinary graph. So we employ the method proposed by Feng \emph{et al.}\cite{HGNN}  to generate hypergrph, i.e. each node as the centroid and its connected nodes form a hypergraph including the centroid itself. 

\textbf{Experimental Setup}. We also use HGNN as the evaluation backbone and employ the same hyper-parameter search process as in the previous experiment. 
And we conduct controlled experiments with the difference of with and without the plugging of HERALD module.
The hyper-parameters  are  used  the same settings as the illustration stated before. 
For obtaining the hypergraph embedding, we add a summation operator as the permutation invariant  layer at the end of the backbone to readout the node embeddings.

\textbf{Results and Disscussion}. The results of  experiments for hypergraph classification are shown in Table \ref{graph classification}. 
Comparing with the results of vanilla GCN and HGNN, it can be ovserved that the performance of HGNN is better than that of GCN, which demonstrates the meaning of  using hypergraph to work as a more powerful tool on modeling complex irregular relationship. 
One can also see that the proposed HERALD module consistently improves the model performance for all the cases and gains 1.12$\%$ test accuracy improvement on average, 
which further verifies the effectiveness and generalization of the approach.
\vspace{-2mm}

\section{conclusion}
\label{sec:majhead}
We have presented a generic plug-in-play module of HypERgrAph Laplacian aDaptor (HERALD)  for improving the representational power of HGCNNs. 
The module is tailored design for constructing task-aware hypergraph topology. To this end, HERALD generates the parameterized hypergrapph laplacian and involves it into the end-to-end training process of HGCNNs. 
The experiments have shown our method gained remarkable performance on both hypernode and hypergraph tasks, which verifies the effectiveness of the method.



\vfill\pagebreak

\bibliographystyle{IEEEbib} 
\bibliography{strings,refs} 

\end{document}